\PassOptionsToPackage{table,dvipsnames}{xcolor}
\documentclass{article} % For LaTeX2e
\usepackage{iclr2027_conference,times}

\newif\ifarxiv
\arxivtrue       % arXiv version
\ifarxiv
    \iclrfinalcopy
\fi

\usepackage{amsmath,amssymb,amsthm,mathtools}
\usepackage{booktabs}
\usepackage{graphicx}
\usepackage{xcolor}
\usepackage{hyperref}
\usepackage{url}
\hypersetup{hidelinks}

\usepackage{algorithm}
\usepackage{algpseudocode}

\newtheorem{theorem}{Theorem}
\newtheorem{proposition}{Proposition}
\newtheorem{lemma}{Lemma}
\newtheorem{corollary}{Corollary}
\newcommand{\calP}{\mathcal{P}}
\newcommand{\calV}{\mathcal{V}}
\newcommand{\calD}{\mathcal{D}}
\newcommand{\ind}{\mathbb{I}}

\newcommand{\Prob}{\mathbb{P}}
\definecolor{OursRow}{HTML}{EAF2FB}
\definecolor{KeyRed}{HTML}{B2182B}
\newcommand{\key}[1]{\textbf{\color{KeyRed}#1}}

\title{View-Structured Conformal Prediction for 3D Gaussian Splatting}

\author{%
\parbox{0.94\textwidth}{%
\centering
Junzheng Chu$^{1}$,
Bin Pan$^{1,*}$,
Zhenwei Shi$^{2}$
\\[1.5mm]
{\small
$^{1}$School of Statistics and Data Science, LEBPS, KLMDASR,\\
LPMC, AAIS, NITFID, Nankai University
\\[0.8mm]
$^{2}$Department of Aerospace Intelligent Science and Technology,\\
School of Astronautics, Beihang University
\\[1mm]
$^{*}$Corresponding author:
\texttt{panbin@nankai.edu.cn}
}%
}%
}

\begin{document}

\maketitle

\ifarxiv
    \fancyhead[L]{Preprint}  % Do not claim acceptance at ICLR
\fi

\begin{abstract}
3D Gaussian Splatting (3DGS) renders novel views in real time, but an
uncertainty heatmap does not certify that a rendered view meets a certain
prediction coverage.  We treat novel-view synthesis as structured
regression and ask that, with probability at least $1-\alpha$, RGB prediction
boxes cover at least a $1-\beta$ fraction of pixels in a new view.  We propose
View-Structured Conformal Prediction (VSCP).  It splits the pre-calibration
scale into a spatial shape from the renderer and a transferable view-difficulty
factor, which predicts the smallest view-wise multiplier that shape needs.  A
held-out quantile over views (View-CP) then gives finite-sample validity even
when transferring to new scenes.  The same factorization makes the analysis exact: a
conformity score is the ratio of oracle to predicted view difficulty, and
excess width separates into a test-side and a calibration-side term.  Across
13 real scenes, pixel-pooled calibration reaches 89.9\% marginal
pixel coverage but only 61.4\% view-event coverage at a 90\% target, while
View-CP reaches 91.7--92.0\%.  At matched coverage VSCP cuts width by 22.1\%
against a constant scale, and matches a ten-model ensemble's 21.0\% reduction using
only one model per scene and four rather than ten rasterization passes per query. 
VSCP also improves on the closest single-model baseline, the 3DGS-U field, by
4.7 points ($p=0.0225$).  The view predictor transfers from
bounded source families to all nine unbounded Mip-NeRF~360 scenes.  There the
full scale beats the constant scale with 20.7\% width saving on all nine scenes.
It also keeps an 18.3\% saving under a different densification backbone and runs at
216--280 FPS on an RTX~4090.
\end{abstract}

\section{Introduction}
\label{sec:intro}

3D Gaussian Splatting (3DGS) makes high-resolution novel-view synthesis
practical in real time~\citep{kerbl2023gs,ren2026fastgs}.  Yet a rendering can
look plausible exactly where it is wrong, and recent methods therefore render
posterior variance, visibility, information, or residual-derived uncertainty
\citep{goli2024bayes,wu2026horseshoe,galappaththige2026gsu,xue2026gavis}.
Such fields are evaluated mainly by AUSE and rank correlation, which read only
the ordering of pixels.  An ordering has no units: it never says how wide an
interval must be, or how often it contains the unknown image.  We find two
scales with identical per-view AUSE and Spearman whose conformal width savings
differ by 12.2 points: our full scale and its own spatial factor.

The statistical unit matters just as much.  A rendered image holds millions of
dependent pixels, but a user looks at one view.  Two methods can both miss 10\%
of all pixels: one misses 10\% in every view, the other misses 20\% in half the
views and none in the rest.  Pooled coverage is the same, yet only the first
gives every view 90\% within-view coverage.  We therefore ask a view-level
question: with probability at least $1-\alpha$, does a new rendering cover at
least a $1-\beta$ fraction of its pixels?

Conformal prediction can certify this event once each held-out view has a
score.  Validity alone does not make the boxes useful: a constant scale is
valid but far too wide in easy views and in well-observed regions.  The real
problem is efficiency under the spatial and angular structure of rendering.  We
address it with the scale
\begin{equation}
  s_{vp}=a(z_v)\,\widetilde b_{vp},
  \label{eq:factorization-intro}
\end{equation}
where $z_v$ is a label-free descriptor of view $v$, $\widetilde b$ is the
relative spatial variation inside that view, and $a$ is its overall difficulty.
We fit $\widetilde b$ from training residuals on view-dependent Gaussian
primitives.  We learn $a$ on source scenes from label-free camera, rendering,
exposure, and directional-support features.

\begin{figure}[t]
  \centering
  \includegraphics[width=\linewidth]{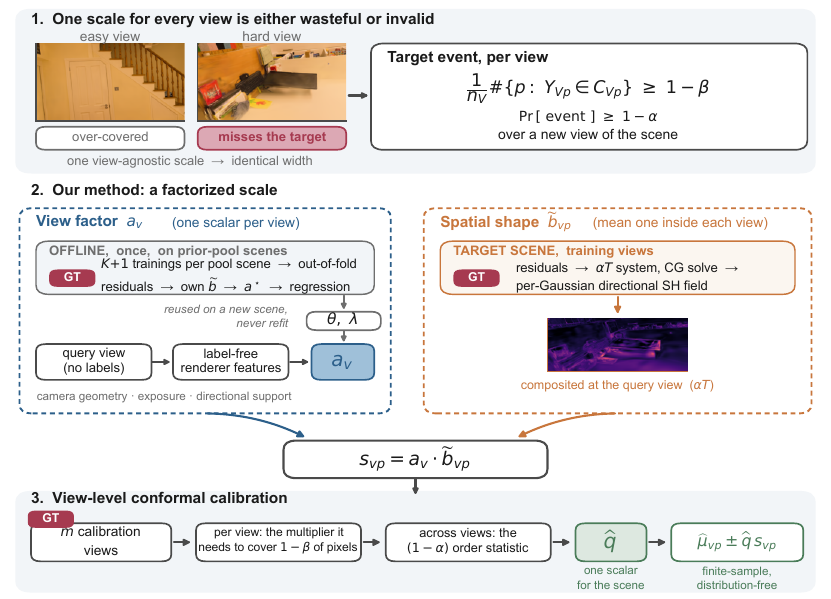}
  \caption{Overview of VSCP.  A single scale can
  waste width on an easy view and miss the within-view target on a hard view.
  We factor the scale into a transferable view factor $a_v$ and a target-scene
  spatial shape $\widetilde b_{vp}$.  Only the held-out calibration views enter
  the nested order statistics.  They produce one multiplier $\widehat q$ and
  finite-sample valid RGB boxes for a new view.}
  \label{fig:overview}
\end{figure}

The regression target for $a$ is not an arbitrary error summary.  For a fixed
spatial shape $\widetilde b$, we derive the smallest multiplier
\begin{equation}
  a_v^\star(\widetilde b)
  =Q_{1-\beta,p}\!\left(r_{vp}/\widetilde b_{vp}\right)
  \label{eq:oracle-intro}
\end{equation}
that lets the view reach its required within-view coverage.  This turns
view-difficulty prediction into a direct efficiency problem.  Since $a$ enters
only the width, it transfers to a new scene, or a new dataset family, without
refitting.  Figure~\ref{fig:overview} summarizes our method.

This form also makes the analysis exact.  A view's conformity score is exactly
$a_v^\star/a_v$, the ratio of oracle to used view difficulty.  The width ratio
against the oracle is then an identity, not a bound: the test view's own error
over a low order statistic of the calibration errors.  One underestimated
calibration view therefore widens every interval, and average regression
accuracy is the wrong efficiency diagnostic.

Our contributions are as follows.  (1) We state reliable 3DGS rendering as a
high-probability within-view RGB coverage event, and build finite-sample valid
boxes that calibrate on views rather than pixels; pixel pooling does not
control this event at any level.  (2) We build a renderer-structured scale over
views and pixels, derive the risk-optimal view target for a fixed spatial
shape, and prove an exact excess-width decomposition in the log view-factor
error.  (3) Its view factor is learned once on source scenes and needs only
label-free features at test time.  It matches a ten-model ensemble scale at 1/10
of the training cost, and beats the closest single-model baseline.  It
transfers to an unseen dataset family and to a different backbone, and exposes
a blind spot in per-view ranking metrics.

\section{Related Work}
\label{sec:related}

\paragraph{Uncertainty for radiance fields and 3DGS.}
Post-hoc Laplace and Fisher methods quantify uncertainty or information in
pretrained radiance fields~\citep{goli2024bayes,jiang2024fisherrf}, and
variational or Bayesian 3DGS models parameter and predictive uncertainty
directly~\citep{li2024vimc,wu2026horseshoe}.  Related signals drive pruning,
active view selection, dynamic reconstruction, and RGB-D
mapping~\citep{hanson2025pup,guo2026usplat4d,tran2026varsplat}.  Closest to us,
3DGS-U fits per-primitive photometric uncertainty from reconstruction residuals
\citep{galappaththige2026gsu}, GAVIS builds an anisotropic visibility field
\citep{xue2026gavis}, and rendering-aware Bayesian 3DGS reports posterior
intervals and their calibration error~\citep{jia2026rendering}.  We reuse the
residual-attribution idea of 3DGS-U as our within-view shape and evaluate the
GAVIS field under a shared rasterizer and calibration layer.  Our target is a
different object: a distribution-free finite-sample guarantee for an explicit
view event, with a learned cross-scene view factor.  We do not claim that
prediction intervals or directional uncertainty are new.

\paragraph{Structured and image-valued conformal prediction.}
Distribution-free image-to-image regression builds simultaneous pixel intervals
for image-valued responses~\citep{angelopoulos2022image}, and conformalized
quantile regression learns heterogeneous scalar intervals~\citep{romano2019cqr}.
Conformal risk control extends split conformal prediction to monotone losses and
quantile risk~\citep{angelopoulos2024crc}.  Conformal structured prediction uses
task structure for large output sets~\citep{zhang2025structured}, and Kandinsky
calibration groups similar pixels~\citep{brunekreef2024kandinsky}.  Our view
event is a task-specific quantile-risk construction, so the new part is the
renderer-structured efficiency model, not the conformal rank argument.

Input-adaptive conformal scales have been learned by image-specific threshold
regression~\citep{luo2026coat}, approximate conditional predictive
distributions~\citep{plassier2025probcp}, and latent-domain weighting
\citep{kong2026moecp}.  We keep an exact marginal-over-views guarantee and use
spatial and angular structure only to reduce width.  Our excess-width result
complements general non-asymptotic analyses of conformalized
regression~\citep{yao2026efficiency}: a multiplicative view factor gives a
simple bound for the rendering event.

3DGS already uses view-dependent spherical harmonics for color, and 6DGS
couples spatial and angular coordinates more explicitly~\citep{gao20256dgs}.
We do not change the renderer.  We use view direction twice for uncertainty.  A
spherical-harmonic residual field says where an interval should widen inside a
view.  Directional observation support helps predict how much the whole view
should widen.

\section{Problem Setup: View-Structured Coverage}
\label{sec:setup}

\subsection{3DGS as functional regression}

Fix a scene and a trained 3DGS renderer.  A view $v$ contains camera extrinsics,
intrinsics, and the associated image-formation conditions.  The renderer maps
this view to an RGB image
$\widehat\mu_v=\{\widehat\mu_{vp}\in[0,1]^3:p\in\calP_v\}$, while the response is
the unknown image $Y_v=\{Y_{vp}\}_{p\in\calP_v}$.  This is a functional
regression problem: one covariate $v$ produces a structured, image-valued
response.  Pixels may be arbitrarily dependent; the statistical observations
used by conformal calibration are views.

Let $w_{vpi}=\alpha_{vpi}T_{vpi}$ be the alpha-transmittance contribution of
Gaussian $i$ to pixel $p$.  Suppressing background terms, the point rendering
has the form
\begin{equation}
  \widehat\mu_{vp}=\sum_i w_{vpi}\,c_{vi},
  \label{eq:gs-render}
\end{equation}
where $c_{vi}$ is the spherical-harmonic color of Gaussian $i$ evaluated for
camera $v$.  Equation~\eqref{eq:gs-render} is used only to construct the scale;
the proposed calibration does not modify the 3DGS point predictor.

\subsection{RGB boxes and the view event}

We use the scalar RGB residual
\begin{equation}
  r_{vp}=\lVert Y_{vp}-\widehat\mu_{vp}\rVert_\infty.
  \label{eq:linf-residual}
\end{equation}
For a positive scale $s_{vp}$ and multiplier $q$, the output set is the
axis-aligned RGB box
\begin{equation}
  C_{vp}(q,s)=
  \prod_{c=1}^{3}
  [\widehat\mu_{vpc}-qs_{vp},\ \widehat\mu_{vpc}+qs_{vp}].
  \label{eq:rgb-box}
\end{equation}
The sup-norm makes $Y_{vp}\in C_{vp}$ exactly equivalent to
$r_{vp}\le qs_{vp}$.  We evaluate unclipped half-width $qs_{vp}$; intersecting
the box with $[0,1]^3$ preserves containment but changes the width functional.

For user parameters $\alpha,\beta\in(0,1)$, define
\begin{equation}
  E_v(q,s)=\left\{
  \frac{1}{n_v}\sum_{p\in\calP_v}
  \ind\{Y_{vp}\in C_{vp}(q,s)\}\ge 1-\beta
  \right\},\qquad n_v=|\calP_v|.
  \label{eq:view-event}
\end{equation}
Our target is
\begin{equation}
  \Prob\{E_V(\widehat q,s)\}\ge 1-\alpha
  \label{eq:target}
\end{equation}
for a new view $V$ exchangeable with the calibration views.  Thus $1-\beta$ is
the required fraction of covered pixels inside a successful view, while
$1-\alpha$ is the probability that a new view is successful.  This is marginal
over views, not conditional coverage for every fixed camera pose.

\section{View-Structured Conformal Prediction}
\label{sec:method}

We call the complete method VSCP: a factorized scale under view-level conformal
calibration (View-CP), the same calibration layer we apply to every baseline.

\subsection{Data roles and factorized scale}

The method uses four disjoint roles.  Scene training views fit the target-scene
3DGS and its residual field.  Source scenes learn a transferable view predictor
and select fixed shrinkage parameters.  Target calibration views are used only
for the final conformal quantile.  Target test views are used only for
evaluation.  This separation is essential: using target calibration labels to
select features or shrinkage would invalidate the stated split-conformal
guarantee.

The scale before conformal calibration is
\begin{align}
 s_{vp} &= a_v\,\widetilde b_{vp},
 \label{eq:scale}\\
 a_v &= (1-\lambda_a)+\lambda_a\widehat a(z_v),
 \qquad
 \widetilde b_{vp}=(1-\lambda_b)+\lambda_b
 \frac{b_{vp}}{\overline b_v},
 \label{eq:shrinkage}
\end{align}
where $\widehat a$ is the view predictor of Section~\ref{sec:view-factor},
$\overline b_v=n_v^{-1}\sum_p b_{vp}$, and
$\lambda_a,\lambda_b\in[0,1]$ are fixed using source data.  We normalize the
positive raw view predictions to have mean one over the unlabeled query batch,
and Equation~\eqref{eq:shrinkage} gives $n_v^{-1}\sum_p\widetilde b_{vp}=1$,
resolving the multiplicative non-identifiability between $a$ and $b$.  Both
choices fix a unit and nothing else: by Lemma~\ref{lem:gauge} a global positive
constant leaves every calibrated interval unchanged.

\subsection{Renderer-derived spatial shape}
\label{sec:spatial}

Let $e_{tp}=\lVert Y_{tp}-\widehat\mu_{tp}\rVert_\infty$ be a residual on a
3DGS training view $t$.  Following the residual-attribution construction of
3DGS-U~\citep{galappaththige2026gsu}, each Gaussian carries a spherical-harmonic
residual field $g_i(d)=c_i^\top\phi(d)$.  For view $t$, define the linear
rendering operator
\begin{equation}
  (\mathcal A_t C)_p=\sum_i w_{tpi}\,c_i^\top\phi(d_{ti}),
  \label{eq:uncertainty-render}
\end{equation}
where $d_{ti}$ is the unit vector from Gaussian $i$ to camera $t$.  We fit
coefficients using the target scene's training residuals:
\begin{equation}
  \widehat C=\arg\min_C
  \sum_{t\in\calD^{\mathrm{GS}}}\lVert\mathcal A_tC-e_t\rVert_2^2
  +\tau\lVert C-C_{\mathrm{prior}}\rVert_2^2,
  \label{eq:spatial-fit}
\end{equation}
where $\tau>0$ is a ridge penalty and $C_{\mathrm{prior}}$ represents a constant
high-uncertainty field.  We use
conjugate gradients with exact forward/adjoint rasterization operators.  At a
query view, $b_{vp}=\max\{(\mathcal A_v\widehat C)_p,\epsilon_0\}$ is rendered like
an additional color channel.  In VSCP $b$ is a relative shape, not a
calibrated standard deviation; its overall scale is removed in
Equation~\eqref{eq:shrinkage}.

\subsection{Angularly aware, risk-derived view factor}
\label{sec:view-factor}

The view descriptor $z_v$ contains only quantities available without query
labels.  Camera features measure nearest-neighbor distance, angular distance,
and local density relative to training cameras.  Render features summarize
accumulated alpha, depth, and RGB gradient.  Most importantly, the renderer
provides per-Gaussian observation support
\begin{equation}
 E_i=\sum_{t,p}w_{tpi},\qquad
 D_i=\sum_t\Bigl(\sum_p w_{tpi}\Bigr)d_{ti}.
 \label{eq:support}
\end{equation}
For query direction $d_{vi}$, we render summaries of exposure $\log(1+E_i)$,
angular extrapolation
$\arccos\!\bigl(d_{vi}^\top D_i/\lVert D_i\rVert\bigr)$, and directional spread
$1-\lVert D_i\rVert/E_i$, summarizing each map by its mean and lower quantiles.
With the camera and render features and two view-level log-means of the spatial
field, this gives a 24-dimensional descriptor.  It explicitly models whether visible Gaussians
were observed, from which directions, and with what directional concentration.

For a fixed $\widetilde b$, the regression target is
\begin{equation}
 a_v^\star(\widetilde b)
 =Q_{1-\beta,p}\!\left(r_{vp}/\widetilde b_{vp}\right),
 \label{eq:oracle-factor}
\end{equation}
where $Q_{1-\beta,p}$ takes the $\lceil n_v(1-\beta)\rceil$-th smallest value
over the pixels $p\in\calP_v$.  It is built from $r_{vp}$, so it needs labels
and exists only on source scenes.  We fit a ridge-regularized log-linear
predictor
\begin{equation}
  \log\widehat a(z_v)=\theta^\top
  \frac{z_v-\mu_z}{\sigma_z}+c,
  \label{eq:view-regression}
\end{equation}
standardizing features and centering targets within each source scene.  The
ridge penalty is selected by leaving out source scenes.  This intentionally
small model makes the angular signals auditable and is appropriate for the
number of source views.  Source-scene meta-training is amortized; a target scene
needs no fold models and does not update $\theta$.

\subsection{View-level conformal calibration}
\label{sec:viewcp}

For any fixed positive scale $s$, define the score of calibration view $v$ as
\begin{equation}
 R_v(s)=Q_{1-\beta,p}\!\left(r_{vp}/s_{vp}\right).
 \label{eq:view-score}
\end{equation}
Given $m$ calibration views, set
\begin{equation}
 k_\alpha=\lceil(m+1)(1-\alpha)\rceil,
 \qquad
 \widehat q=
 \begin{cases}
   Q_{k_\alpha}\!\left(\{R_1,\ldots,R_m\}\right), & k_\alpha\le m,\\
   +\infty, & k_\alpha>m,
 \end{cases}
 \label{eq:outer-quantile}
\end{equation}
where $Q_{k_\alpha}$ takes the $k_\alpha$-th smallest of the $m$ calibration
scores.  For a new view $V$ we render the center $\widehat\mu_V$, form its
label-free scale, and return
\begin{equation}
  C_{Vp}=\prod_{c=1}^{3}
  \bigl[\widehat\mu_{Vpc}-\widehat q\,a_V\widetilde b_{Vp},\
        \widehat\mu_{Vpc}+\widehat q\,a_V\widetilde b_{Vp}\bigr].
  \label{eq:deployed-interval}
\end{equation}
The half-width varies with the pixel through $\widetilde b_{Vp}$ and with the
view through $a_V$, while $\widehat q$ is a single scalar for the scene.
Algorithm~\ref{alg:vscp} collects the whole procedure, with the data each step
is allowed to read.

\subsection{Rendering cost}

The RGB prediction uses one rasterization.  Seven per-Gaussian auxiliary
scalars are packed into the three RGB channels of three additional
rasterizations.  Channels are composited independently with the same
$\alpha T$ weights, so packing is an algebraic reorganization rather than an
approximation.  The complete query therefore uses one backbone and four
rasterization passes.  Spatial-field fitting and support accumulation are
one-time post-training operations.

\begin{algorithm}[t]
\caption{View-Structured Conformal Prediction (VSCP).}
\label{alg:vscp}
\begin{algorithmic}[1]
\Statex \textbf{Phase 1: meta-training on source scenes (offline, once)}
\For{each source scene}
  \State fit its 3DGS and its residual field $\widehat C$ on that scene's
         training views; accumulate $E_i,D_i$
  \For{each fold $k$, and each training view $v$ held out of fold $k$}
    \State $z_v\gets$ descriptor of $v$ built from the cameras and support of
           the \emph{other} folds
    \State $\widetilde b_{v}\gets$ \emph{that scene's own} shape at $v$
           \Comment{Equation~\eqref{eq:shrinkage}}
    \State $a_v^\star\gets Q_{1-\beta,p}(r_{vp}/\widetilde b_{vp})$
           \Comment{needs a label; source scenes only}
  \EndFor
\EndFor
\State centre $\log a^\star$ within each scene, pool, fit $\theta$, and select
       $\lambda_a,\lambda_b$
\Statex \emph{$\theta,\lambda_a,\lambda_b$ are frozen from here on and are the
        only objects that cross the scene boundary.}
\Statex
\Statex \textbf{Phase 2: target scene, training views}
\State fit the 3DGS $\widehat\mu$ and the residual field $\widehat C$;
       accumulate $E_i,D_i$
\State compute the feature mean and scale $\mu_z,\sigma_z$ on these views
\Statex
\Statex \textbf{Phase 3: target scene, whole query batch $\calV$, no labels}
\For{every held-out view $v\in\calV$}
  \State render $\widehat\mu_v$ and $b_{vp}$, and form $\widetilde b_{vp}$
         \Comment{Equation~\eqref{eq:shrinkage}}
  \State build $z_v$ and evaluate $\widehat a(z_v)$
         \Comment{Equation~\eqref{eq:view-regression}}
\EndFor
\State \label{alg:batch} normalize $\widehat a$ to mean one \textbf{over all of
       $\calV$}; set $a_v$ and $s_{vp}=a_v\widetilde b_{vp}$
\Statex
\Statex \textbf{Phase 4: target scene, calibration and output}
\State draw $m$ calibration views from $\calV$; the remainder are test views
\For{each calibration view $v$}
  \State \label{alg:score} $R_v\gets Q_{1-\beta,p}(r_{vp}/s_{vp})$
         \Comment{the only use of a target label}
\EndFor
\State $\widehat q\gets Q_{k_\alpha}(\{R_1,\dots,R_m\})$
       \Comment{Equation~\eqref{eq:outer-quantile}}
\For{each test view $V$}
  \State \textbf{output} $\widehat\mu_{Vp}\pm\widehat q\,a_V\widetilde b_{Vp}$
         \Comment{Equation~\eqref{eq:deployed-interval}}
\EndFor
\end{algorithmic}
\end{algorithm}

\section{Validity and Efficiency}
\label{sec:theory}

We condition on the trained renderer, the source scenes, and every choice used
to build the scale.  The exchangeable units are the labeled calibration and
test views, not the pixels inside them.  We build the scales from the unlabeled
covariates of the whole query batch, before any calibration/test split.  This
construction is permutation equivariant, so the scores stay exchangeable
(Proposition~\ref{prop:equivariance}, Appendix~\ref{app:proofs}).

Everything below rests on one simple fact.  Because $s_{vp}>0$, the within-view
coverage fraction is the empirical distribution function of
$\{r_{vp}/s_{vp}\}$.  The view event in Equation~\eqref{eq:view-event}
therefore holds at multiplier $q$ if and only if $q\ge R_v(s)$
(Lemma~\ref{lem:event}, Appendix~\ref{app:proofs}).  This is also why both
quantiles must be discrete order statistics: an interpolated quantile breaks
the equivalence.

\begin{theorem}[Finite-sample view-event validity]
\label{thm:validity}
Suppose $s$ is fixed before target calibration labels are observed and the $m$
calibration scores together with the new-view score are exchangeable.  Then the
boxes calibrated by Equation~\eqref{eq:outer-quantile} satisfy
\begin{equation}
  \Prob\left\{\frac{1}{n_V}\sum_p
  \ind\{Y_{Vp}\in C_{Vp}(\widehat q,s)\}\ge1-\beta\right\}
  \ge1-\alpha.
  \label{eq:validity}
\end{equation}
If scores have no ties and $k_\alpha\le m$, the probability is at most
$1-\alpha+1/(m+1)$.
\end{theorem}

The proof is the standard split-conformal rank argument after
Lemma~\ref{lem:event} (Appendix~\ref{app:proofs}).  We next state the
consequence that licenses transfer.

\begin{corollary}[Transfer]
\label{cor:transfer}
  Let $\mathcal F$ collect every object fixed before target calibration labels
  are observed, including the trained renderer, source data, and fitted scale
  construction.  If the target calibration and test views are
exchangeable given $\mathcal F$, then Equation~\eqref{eq:validity} holds for
every $\mathcal F$.  Mismatch between $\mathcal F$ and the target can change
the interval width, but not the coverage guarantee.
\end{corollary}

We now give the two results that govern width.  Because $\widetilde b$
has within-view mean one, the mean half-width of view $i$ is
$W_i=\widehat q\,n_i^{-1}\sum_p s_{ip}=\widehat q\,a_i$.

\begin{proposition}[Oracle view factor and score identity]
\label{prop:oracle}
Fix a positive $\widetilde b$.  For every view $v$ and every scalar $c>0$,
\begin{equation}
  R_v(c\,\widetilde b)=\frac{a_v^\star(\widetilde b)}{c},
  \label{eq:ratio-identity}
\end{equation}
and $a_v^\star(\widetilde b)$ is the smallest scalar in the family
$\{a\widetilde b:a>0\}$ that attains the within-view event at $q=1$.
\end{proposition}

A conformity score is therefore a ratio of oracle to used view difficulty.
This makes $\log a_v^\star$ the natural regression target.  Appendix
\ref{app:additional-theory} adds its identification under a separable residual
model (Proposition~\ref{prop:identification}), and a uniform error bound.

\begin{proposition}[Exact excess-width decomposition]
\label{prop:decomposition}
Assume $k_\alpha\le m$.  Let $\delta_v=\log(a_v/a_v^\star)$ be the log error of
the deployed post-shrinkage factor of Equation~\eqref{eq:shrinkage}, and let
$\delta_{(1)}\le\cdots\le\delta_{(m)}$ be the ordered calibration errors.  Set
$j_\alpha=\lfloor(m+1)\alpha\rfloor$.  For every test view $i$,
\begin{equation}
 \widehat q=e^{-\delta_{(j_\alpha)}},\qquad
 \frac{W_i(a)}{W_i(a^\star)}
 =e^{\delta_i-\delta_{(j_\alpha)}}.
 \label{eq:exact-width}
\end{equation}
Consequently, with $\overline{(\cdot)}$ the average over the test views of a
fixed split,
\begin{equation}
 \frac{\overline W(a)}{\overline W(a^\star)}
 =\underbrace{\frac{\overline{a^\star e^{\delta}}}{\overline{a^\star}}}_{\text{test-side dispersion}}
 \underbrace{e^{-\delta_{(j_\alpha)}}}_{\text{calibration-side lower tail}}.
 \label{eq:aggregate-decomposition}
\end{equation}
\end{proposition}

Adding a constant to every $\delta_v$ leaves
Equation~\eqref{eq:exact-width} unchanged, which is Lemma~\ref{lem:gauge}:
global multiplicative bias cancels exactly.  At the smallest usable calibration
size for $\alpha=0.1$ we have $m=9$ and $j_\alpha=1$, so one underestimated
calibration view sets the width inflation.  Appendix
\ref{app:additional-theory} also gives an exact realizable-gain identity, and
Appendix~\ref{app:additional-experiments} audits all three identities
numerically.

\section{Experiments}
\label{sec:experiments}

\subsection{Setup, protocols, and metrics}

We evaluate 13 real scenes from Tanks \& Temples~\citep{knapitsch2017tanks}
(truck, train), Deep Blending~\citep{hedman2018deepblending} (drjohnson,
playroom), and Mip-NeRF~360~\citep{barron2022mipnerf360} (nine indoor and
outdoor scenes).  Full-benchmark experiments use FastGS~\citep{ren2026fastgs}
trained for 30k iterations.  A controlled second-backbone study replaces only
FastGS densification and pruning by vanilla 3DGS rules on three scenes.
Everything else is held fixed.

The default event is $(\alpha,\beta)=(0.1,0.1)$.  We repeat each
calibration/test partition 200 times (100 for the native-calibration audit).
These splits reuse the same views and quantify split randomization, not
new-scene uncertainty.  We therefore use scenes as the sampling unit for
paired bootstrap intervals, win counts, and exact two-sided sign tests.  We
report pooled-pixel coverage, view-event coverage, and mean unclipped
half-width.  Relative saving is measured against a constant-scale conformal
predictor under the same point center, calibration views, and event target.

Same-center baselines are a constant scale, the residual field of
3DGS-U~\citep{galappaththige2026gsu}, the anisotropic visibility field of
GAVIS~\citep{xue2026gavis} rendered in our rasterizer, and the pixel standard
deviation of a ten-member deep ensemble~\citep{lakshminarayanan2017deep}.  Each
baseline field is scaled by one label-free scene constant, which keeps its
cross-view magnitude.  Per-view normalization is used only for our
$\widetilde b$, where $a_v$ carries that magnitude
(Appendix~\ref{app:implementation}).  All shrinkage and GAVIS concentration
values come from source scenes, and no target test label is used.  Systems with
their own predictive center are reported separately, since a new center and a
new scale answer different questions.

\subsection{Does calibration control the right event?}

\begin{figure}[t]
  \centering
  \includegraphics[width=\linewidth]{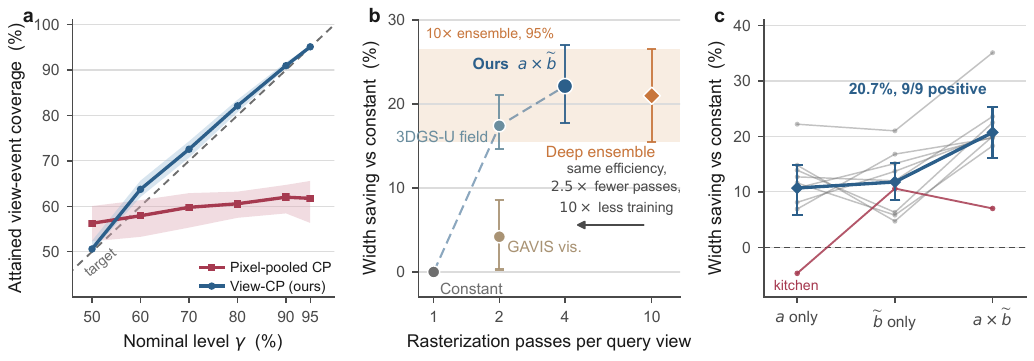}
  \caption{Validity, cost, and strict transfer.  \textbf{(a)} View-CP tracks
  the requested view-event level, while pixel-pooled CP stays near 60\% despite
  89.9\% marginal pixel coverage at the 90\% target.  Curves are scene means;
  bands are interquartile ranges.  \textbf{(b)} Across 13 scenes, VSCP matches
  the ten-model ensemble scale (shaded 95\% interval) using one model and four
  query passes instead of ten models and ten passes.  \textbf{(c)} Under strict family holdout, the
  two factors remain complementary and their product improves over the
  constant scale on all nine target scenes.  Diamonds show means with 95\%
  scene-bootstrap intervals.}
  \label{fig:validity-efficiency}
\end{figure}

Pixel pooling and View-CP read the same scale map differently: the former pools
all calibration pixels, whereas the latter applies
Equations~\eqref{eq:view-score}--\eqref{eq:outer-quantile}.  At the 90\% target,
pixel pooling attains 89.9\% marginal pixel coverage but only 61.4\% view-event
coverage; View-CP attains 91.7--92.0\% across three very different scale maps
(Appendix Table~\ref{tab:validity}).  The failure persists across operating
points: pooled calibration stays nearly flat while View-CP tracks the requested
level (Figure~\ref{fig:validity-efficiency}(a)).

Pixel pooling controls the mean of within-view coverage, not the probability
that a view clears a required fraction.  That one calibration layer serves
three very different scales is Theorem~\ref{thm:validity} applied three times:
the scale sets width and the view order statistic sets validity.  We do not claim
that every scene exceeds 90\% on its own point estimate.  The observed range
over scale variants is 89.4--93.8\%, in line with finite-view split noise.

\subsection{Matched-coverage scale efficiency}

Table~\ref{tab:panel-a} fixes the FastGS RGB center, $L_\infty$ residuals,
calibration views, and View-CP layer.  Only the positive scale map changes.
This is the clean comparison of uncertainty structure.

\begin{table}[t]
\caption{Same-center, same-View-CP comparison over 13 scenes.  Models / passes
denote trained models per scene and rasterizations per query view.  Savings are
relative to the constant scale at matched coverage; intervals bootstrap
scenes.  The GAVIS row uses its visibility field in our shared rasterizer.}
\label{tab:panel-a}
\centering
\begin{tabular}{lcccc}
\toprule
Scale & Models / passes & Saving (\%) & 95\% interval & Event cov. (\%) \\
\midrule
Constant & $1\times$ / 1 & 0.0 & -- & 91.7 \\
3DGS-U spatial field & $1\times$ / 2 & 17.4 & [14.6, 21.0] & 91.6 \\
GAVIS vis. field & $1\times$ / 2 & 4.2 & [0.3, 8.5] & 91.7 \\
Ensemble std. & $10\times$ / 10 & 21.0 & [15.5, 26.4] & 91.4 \\
\rowcolor{OursRow}
Ours (VSCP) $a\times\widetilde b$ & $1\times$ / 4 & \key{22.1} & [17.7, 27.1] & 91.5 \\
\bottomrule
\end{tabular}
\end{table}

The paired gap to the ensemble scale is 1.2 points [-4.0, 6.6], with 7/13 wins
and $p=1.0$.  We therefore claim no efficiency advantage over it.  Our claim is
about cost: one trained model per scene and four passes per query view, against
ten models and ten passes (Figure~\ref{fig:validity-efficiency}(b)).  The
ensemble is also the strongest baseline on the metric we criticize, since it
has the best per-view AUSE and Spearman of every scale we tried (Appendix
Table~\ref{tab:ranking}).  Against
3DGS-U, the closest single-model baseline, the gain is 4.7 points [1.6, 7.8],
with 11/13 wins and $p=0.0225$.  This is the main result of the same-center
comparison.  The table studies mechanisms, not strict transfer: eight of the 13
scenes also belong to the source pool used to select shrinkage.

The GAVIS sanity check confirms directional sensitivity, but interleaved test
views are about as visible as training views.  Visibility therefore does not
identify the held-out views with large photometric residuals
(Appendix~\ref{app:gavis}).

\subsection{Cross-family transfer}

Trained only on two Tanks \& Temples and two Deep Blending scenes, VSCP
transfers without refitting to all nine Mip-NeRF~360 scenes.  It saves 20.7\%
against the constant scale on all nine ($p=0.0039$) and matches the
source-selected ensemble scale at 20.0\%, although that scale requires ten
target-specific models.  A same-size source pool containing Mip-NeRF~360
reaches 21.9\%, only 1.2 points higher.  Corollary~\ref{cor:transfer} keeps
target-scene View-CP valid under such efficiency shifts.  Same-family and
predictive-center controls are in Appendix Table~\ref{tab:transfer}.

\subsection{Factorization, metric blind spots, and theory audit}

\begin{table}[t]
\caption{Factorization analysis over 13 scenes.  $\overline b_v$ is the
per-view mean of the raw spatial field.  Scene-global $b$ retains this
cross-view magnitude but uses one shrinkage parameter for the full field.}
\label{tab:ablation}
\centering
\begin{tabular}{lc}
\toprule
Scale construction & Saving (\%) \\
\midrule
Within-view shape $\widetilde b$ only & 9.9 \\
Raw-field magnitude $\overline b_v$ only & 12.6 \\
Learned view factor $a$ only & 13.6 \\
Scene-global $b$, one shrinkage (the 3DGS-U row of
Table~\ref{tab:panel-a}) & 17.4 \\
Full $a\times\widetilde b$ (separate shrinkage) & \key{22.1} \\
\bottomrule
\end{tabular}
\end{table}

Table~\ref{tab:ablation} separates the value of the view signal from the value
of the structure.  The raw field magnitude already carries view information.
Our learned factor is closer to oracle view difficulty than that magnitude
(mean Spearman 0.719 against 0.677), and is a little better on its own
(13.6\% against 12.6\%).  The larger gain comes from the structure.
Scene-global $b$ uses one shrinkage parameter for cross-view magnitude and
within-view shape together, and saves 17.4\%.  Estimating and shrinking the two
parts separately saves 22.1\%.

Figure~\ref{fig:validity-efficiency}(c) shows that the two factors stay
complementary under strict holdout, with per-scene values in Appendix
Table~\ref{tab:per-scene}.  The oracle factor reaches 39.0\% there, so
better view-difficulty prediction is still the main room for improvement.  A
target built from one deployed model also matches an out-of-fold target that
needs 11 trainings, at one eleventh of the source cost
(Appendix~\ref{app:factor-diagnostics}).

Common ranking metrics cannot see this.  By Lemma~\ref{lem:nullspace} the view
factor lies in the null space of per-view AUSE and Spearman correlation:
$a\widetilde b$ and $\widetilde b$ score identically on both, yet save 22.1\%
and 9.9\% (Appendix Table~\ref{tab:ranking}).

\begin{figure}[t]
  \centering
  \includegraphics[width=\linewidth]{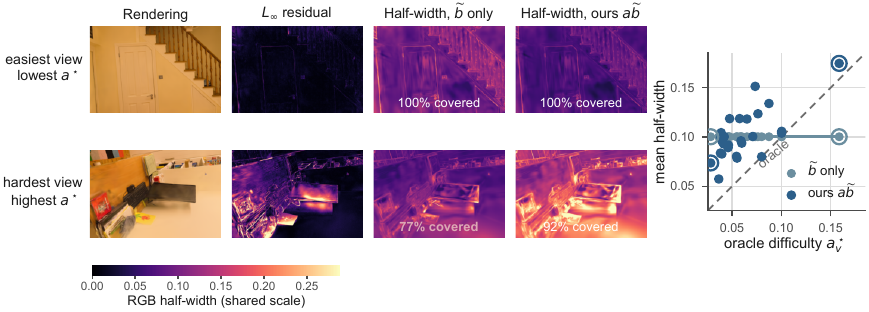}
  \caption{The view factor is invisible within a view and decisive across
  views.  In the preselected playroom scene, rows show the easiest and hardest
  held-out views by oracle difficulty.  Multiplication by a positive view
  factor preserves the pixel ranking, so the two scale maps in each row should
  look alike.  Across views it raises hard-view coverage from 77\% to 92\% at a
  90\% target.  Right: mean width over all 29 views; $\widetilde b$ alone is
  flat by construction.}
  \label{fig:nullspace}
\end{figure}

The numerical audit further shows that the lower tail of calibration error,
rather than average regression error, dominates excess width
(Appendix~\ref{app:factor-diagnostics}).

\subsection{Robustness, labels, and query cost}

Without refitting, VSCP retains an 18.3\% saving after replacing FastGS density
control with vanilla 3DGS rules, despite 5.6--7.7$\times$ more Gaussians.  A
finite View-CP threshold requires $m\ge9$ at $\alpha=0.1$ and $m\ge19$ at
$\alpha=0.05$; below this limit the valid set is unbounded.  Exact channel
packing uses four passes and runs at 216--280 FPS.  Target sensitivity,
calibration-size sweeps, backbone results, and timing details are in
Appendix~\ref{app:additional-experiments}.

% Limitations moved to Appendix E to free main-text space.  The text below is
% kept verbatim so the section can be restored by uncommenting it and deleting
% Appendix E.  ICLR does not require a Limitations section in the main text,
% but reviewers do look for the content, so it is discussed, not dropped.
%
% \section{Limitations}
% \label{sec:limitations}
%
% The guarantee is marginal over exchangeable views, not conditional coverage for
% every camera pose, and ordered camera paths can break exchangeability.  The
% method needs labeled target-scene views that are held out.  Using them to train
% 3DGS instead would need a different calibration argument.  Only one held-out
% family is large enough to be useful, and family is confounded with bounded
% versus unbounded capture.  The 13-scene results use FastGS.  The
% vanilla-densification study covers three scenes and changes only the density
% rule, so it is not the official vanilla implementation.  Our GAVIS row
% evaluates the released visibility field under a shared rasterizer, not its
% Bayesian uncertainty head.  Systems such as Horseshoe Splatting change the
% center and the training objective, so we discuss them rather than place them in
% Table~\ref{tab:panel-a}.  Finally, our sets are axis-aligned RGB boxes with a
% shared radius.  Richer color geometry, dynamic scenes, and time-correlated
% calibration units are out of scope.

\section{Conclusion}

Reliable rendering requires saying what should be covered, at what statistical
unit, and at what cost.  We treat a 3DGS view as a structured regression
response and calibrate the event that most of its pixels are covered.
Renderer-derived spatial and angular structure then makes the boxes efficient.
The conformal quantile protects validity while the factorized scale controls
width.

\bibliography{references}

@article{kerbl2023gs,
  title   = {3D Gaussian Splatting for Real-Time Radiance Field Rendering},
  author  = {Kerbl, Bernhard and Kopanas, Georgios and Leimk{\"u}hler, Thomas and Drettakis, George},
  journal = {ACM Transactions on Graphics},
  volume  = {42},
  number  = {4},
  year    = {2023}
}

@inproceedings{angelopoulos2022image,
  title     = {Image-to-Image Regression with Distribution-Free Uncertainty Quantification and Applications in Imaging},
  author    = {Angelopoulos, Anastasios N. and Kohli, Amit Pal and Bates, Stephen and Jordan, Michael and Malik, Jitendra and Alshaabi, Thayer and Upadhyayula, Srigokul and Romano, Yaniv},
  booktitle = {Proceedings of the 39th International Conference on Machine Learning},
  series    = {Proceedings of Machine Learning Research},
  volume    = {162},
  pages     = {717--730},
  year      = {2022}
}

@inproceedings{angelopoulos2024crc,
  title     = {Conformal Risk Control},
  author    = {Angelopoulos, Anastasios N. and Bates, Stephen and Fisch, Adam and Lei, Lihua and Schuster, Tal},
  booktitle = {International Conference on Learning Representations},
  year      = {2024}
}

@inproceedings{zhang2025structured,
  title     = {Conformal Structured Prediction},
  author    = {Zhang, Botong and Li, Shuo and Bastani, Osbert},
  booktitle = {International Conference on Learning Representations},
  year      = {2025}
}

@inproceedings{brunekreef2024kandinsky,
  title     = {Kandinsky Conformal Prediction: Efficient Calibration of Image Segmentation Algorithms},
  author    = {Brunekreef, Joren and Marcus, Eric and Sheombarsing, Ray and Sonke, Jan-Jakob and Teuwen, Jonas},
  booktitle = {Proceedings of the IEEE/CVF Conference on Computer Vision and Pattern Recognition},
  pages     = {4135--4143},
  year      = {2024}
}

@inproceedings{luo2026coat,
  title     = {Enhancing Image-Conditional Coverage in Segmentation: Adaptive Thresholding via Differentiable Miscoverage Loss},
  author    = {Luo, Rui and Bao, Jie and Su, Xiaoyi and Li, Wen Jung and Cao, Suqun},
  booktitle = {International Conference on Learning Representations},
  year      = {2026}
}

@inproceedings{plassier2025probcp,
  title     = {Probabilistic Conformal Prediction with Approximate Conditional Validity},
  author    = {Plassier, Vincent and Fishkov, Alexander and Guizani, Mohsen and Panov, Maxim and Moulines, Eric},
  booktitle = {International Conference on Learning Representations},
  year      = {2025}
}

@inproceedings{kong2026moecp,
  title     = {Adaptive Conformal Prediction via Mixture-of-Experts Gating Similarity},
  author    = {Kong, Jingsen and Tang, Wenlu and Kong, Dezheng and Kong, Linglong and Yang, Guangren and Jiang, Bei},
  booktitle = {International Conference on Learning Representations},
  year      = {2026}
}

@inproceedings{yao2026efficiency,
  title     = {Non-Asymptotic Analysis of Efficiency in Conformalized Regression},
  author    = {Yao, Yunzhen and He, Lie and Gastpar, Michael},
  booktitle = {International Conference on Learning Representations},
  year      = {2026}
}

@inproceedings{gao20256dgs,
  title     = {6DGS: Enhanced Direction-Aware Gaussian Splatting for Volumetric Rendering},
  author    = {Gao, Zhongpai and Planche, Benjamin and Zheng, Meng and Choudhuri, Anwesa and Chen, Terrence and Wu, Ziyan},
  booktitle = {International Conference on Learning Representations},
  year      = {2025}
}

@inproceedings{wu2026horseshoe,
  title     = {Horseshoe Splatting: Handling Structural Sparsity for Uncertainty-Aware Gaussian-Splatting Radiance Field Rendering},
  author    = {Wu, Feng and Chan, Tsai Hor and Chen, Yihang and Zhu, Lingting and Yin, Guosheng and Yu, Lequan},
  booktitle = {International Conference on Learning Representations},
  year      = {2026}
}

@inproceedings{guo2026usplat4d,
  title     = {Uncertainty Matters in Dynamic Gaussian Splatting for Monocular 4D Reconstruction},
  author    = {Guo, Fengzhi and Hsu, Chih-Chuan and Ding, Sihao and Zhang, Cheng},
  booktitle = {International Conference on Learning Representations},
  year      = {2026}
}

@article{galappaththige2026gsu,
  title   = {Predictive Photometric Uncertainty in Gaussian Splatting for Novel View Synthesis},
  author  = {Galappaththige, Chamuditha Jayanga and Gottwald, Thomas and Stehr, Peter and Heinert, Edgar and S{\"u}nderhauf, Niko and Miller, Dimity and Rottmann, Matthias},
  journal = {arXiv preprint arXiv:2603.22786},
  year    = {2026}
}

@inproceedings{xue2026gavis,
  title     = {Uncertainty-Driven 3D Gaussian Splatting Active Mapping via Anisotropic Visibility Field},
  author    = {Xue, Shangjie and Dill, Jesse and Ahuja, Dhruv and Dellaert, Frank and Tsiotras, Panagiotis and Xu, Danfei},
  booktitle = {Proceedings of the IEEE/CVF Conference on Computer Vision and Pattern Recognition},
  pages     = {5014--5026},
  year      = {2026}
}

@article{jia2026rendering,
  title   = {Rendering-Aware Bayesian 3D Gaussian Splatting with Native Uncertainty and Adaptive Complexity Control},
  author  = {Jia, Gaoxiang and Appia, Vikram and Huang, Junzhou and Wang, Xinlei},
  journal = {arXiv preprint arXiv:2607.05522},
  year    = {2026}
}

@inproceedings{goli2024bayes,
  title     = {Bayes' Rays: Uncertainty Quantification for Neural Radiance Fields},
  author    = {Goli, Lily and Reading, Cody and Sell{\'a}n, Silvia and Jacobson, Alec and Tagliasacchi, Andrea},
  booktitle = {Proceedings of the IEEE/CVF Conference on Computer Vision and Pattern Recognition},
  pages     = {20061--20070},
  year      = {2024}
}

@inproceedings{jiang2024fisherrf,
  title     = {FisherRF: Active View Selection and Mapping with Radiance Fields Using Fisher Information},
  author    = {Jiang, Wen and Lei, Boshu and Daniilidis, Kostas},
  booktitle = {European Conference on Computer Vision},
  year      = {2024}
}

@inproceedings{li2024vimc,
  title     = {Variational Multi-Scale Representation for Estimating Uncertainty in 3D Gaussian Splatting},
  author    = {Li, Ruiqi and Cheung, Yiu-ming},
  booktitle = {Advances in Neural Information Processing Systems},
  year      = {2024}
}

@inproceedings{ren2026fastgs,
  title     = {{FastGS}: Training 3D Gaussian Splatting in 100 Seconds},
  author    = {Ren, Shiwei and Wen, Tianci and Fang, Yongchun and Lu, Biao},
  booktitle = {Proceedings of the IEEE/CVF Conference on Computer Vision and Pattern Recognition},
  pages     = {26094--26103},
  year      = {2026}
}

@inproceedings{lakshminarayanan2017deep,
  title     = {Simple and Scalable Predictive Uncertainty Estimation Using Deep Ensembles},
  author    = {Lakshminarayanan, Balaji and Pritzel, Alexander and Blundell, Charles},
  booktitle = {Advances in Neural Information Processing Systems},
  volume    = {30},
  year      = {2017}
}

@inproceedings{romano2019cqr,
  title     = {Conformalized Quantile Regression},
  author    = {Romano, Yaniv and Patterson, Evan and Cand{\`e}s, Emmanuel J.},
  booktitle = {Advances in Neural Information Processing Systems},
  volume    = {32},
  year      = {2019}
}

@article{knapitsch2017tanks,
  title   = {Tanks and Temples: Benchmarking Large-Scale Scene Reconstruction},
  author  = {Knapitsch, Arno and Park, Jaesik and Zhou, Qian-Yi and Koltun, Vladlen},
  journal = {ACM Transactions on Graphics},
  volume  = {36},
  number  = {4},
  year    = {2017}
}

@article{hedman2018deepblending,
  title   = {Deep Blending for Free-Viewpoint Image-Based Rendering},
  author  = {Hedman, Peter and Philip, Julien and Price, True and Frahm, Jan-Michael and Drettakis, George and Brostow, Gabriel},
  journal = {ACM Transactions on Graphics},
  volume  = {37},
  number  = {6},
  year    = {2018}
}

@inproceedings{barron2022mipnerf360,
  title     = {{Mip-NeRF} 360: Unbounded Anti-Aliased Neural Radiance Fields},
  author    = {Barron, Jonathan T. and Mildenhall, Ben and Verbin, Dor and Srinivasan, Pratul P. and Hedman, Peter},
  booktitle = {Proceedings of the IEEE/CVF Conference on Computer Vision and Pattern Recognition},
  pages     = {5470--5479},
  year      = {2022}
}

@inproceedings{hanson2025pup,
  title     = {{PUP 3D-GS}: Principled Uncertainty Pruning for 3D Gaussian Splatting},
  author    = {Hanson, Alex and Tu, Allen and Singla, Vasu and Jayawardhana, Mayuka and Zwicker, Matthias and Goldstein, Tom},
  booktitle = {Proceedings of the IEEE/CVF Conference on Computer Vision and Pattern Recognition},
  pages     = {5949--5958},
  year      = {2025}
}

@inproceedings{tran2026varsplat,
  title     = {{VarSplat}: Uncertainty-Aware 3D Gaussian Splatting for Robust {RGB-D} {SLAM}},
  author    = {Tran, Anh Thuan and Kosecka, Jana},
  booktitle = {Proceedings of the IEEE/CVF Conference on Computer Vision and Pattern Recognition},
  year      = {2026}
}
\bibliographystyle{iclr2027_conference}

\newpage

\appendix

\section{Additional Efficiency Results}
\label{app:additional-theory}

Section~\ref{sec:theory} treats $\log a_v^\star$ as the regression target
without saying when it recovers real view difficulty.  The following
proposition answers that, and degrades gracefully when the assumption only
holds approximately.

\begin{proposition}[Identification of the oracle view factor]
\label{prop:identification}
Fix the deployed shape $\widetilde b$ and assume the separable model
$r_{vp}=A_v\,\widetilde b_{vp}\,\varepsilon_{vp}$, in which $A_v>0$ is the
difficulty of view $v$ relative to that shape.  Write
$c_v=Q_{1-\beta,p}(\varepsilon_{vp})$ for the
inner order statistic of the view's own noise.  Then
$a_v^\star(\widetilde b)=A_vc_v$, and:
\begin{enumerate}
\item if $c_v\equiv c_\beta$ across views, the oracle factor equals $A_v$ up to
a single global constant, which by Lemma~\ref{lem:gauge} changes no interval;
\item if $|\log c_v-\log c_\beta|\le\eta$ and the deployed factor tracks $A_v$
with centered log error at most $\epsilon$, then
$\lVert\delta\rVert_\infty\le\epsilon+\eta$ and Theorem~\ref{thm:width} gives
$W_i(a)/W_i(a^\star)\le e^{2(\epsilon+\eta)}$.
\end{enumerate}
\end{proposition}

The identity is immediate from positive homogeneity: dividing $r_{vp}$ by
$\widetilde b_{vp}$ leaves $A_v\varepsilon_{vp}$, whose
$\lceil n_v(1-\beta)\rceil$-th order statistic is $A_vc_v$.  Two things are
worth noting.  No pixel-independence assumption is used; what
``view-invariant noise'' must control is only the inner quantile $c_v$ of each
view, not its full noise law.  And because $\widetilde b$ is normalized within
each view, $A_v$ absorbs the per-view magnitude of the underlying spatial
field, which is why that magnitude alone already carries view information in
Table~\ref{tab:ablation}.

\begin{theorem}[Deterministic excess-width bound]
\label{thm:width}
Assume $k_\alpha\le m$.  Fix $\widetilde b$ and suppose the post-shrinkage log
errors $\delta_v=\log(a_v/a_v^\star)$ satisfy
$\lVert\delta\rVert_\infty\le\epsilon$ over all calibration and test views.  For
every test view $i$,
\begin{equation}
  e^{-2\epsilon}
  \le \frac{W_i(a)}{W_i(a^\star)}
  \le e^{2\epsilon}.
  \label{eq:width-bound}
\end{equation}
Hence the multiplicative excess width is at most
$e^{2\epsilon}-1=2\epsilon+O(\epsilon^2)$.
\end{theorem}

By Equation~\eqref{eq:ratio-identity}, calibration scores are
$e^{-\delta_v}$, so their order statistic lies in
$[e^{-\epsilon},e^{\epsilon}]$; multiplying by
$a_i=a_i^\star e^{\delta_i}$ gives the result.  The theorem is
deterministic and needs neither a score-density lower bound nor an asymptotic
calibration approximation.  It also follows directly from the exact
decomposition because
$|\delta_i-\delta_{(j_\alpha)}|\le2\epsilon$.

The same factorization identifies the realizable benefit of view adaptation.
For the oracle factor versus a constant view factor,
\begin{equation}
 \frac{\overline W(a^\star)}{\overline W(1)}
 =\frac{\overline{a^\star}_{\mathrm{test}}}
 {Q_{k_\alpha}\!\left(\{a_v^\star:v\in\mathrm{cal}\}\right)}.
 \label{eq:realizable-gain-main}
\end{equation}
This is an exact finite-sample identity conditional on one calibration/test
split: a perfect view model replaces an upper quantile of view difficulty by
its mean, and absolute scene error cancels.  It predicts larger gains when
view difficulty is more heterogeneous and holds numerically to $6.2\times
10^{-7}$ over all 13 scenes.

\section{Proofs}
\label{app:proofs}

\subsection{Gauge invariance}

\begin{lemma}[Global scale invariance]
\label{lem:gauge}
Let $s$ be a positive scale and $c>0$ a constant.  Then $R_v(cs)=R_v(s)/c$ for
every view, hence $\widehat q(cs)=\widehat q(s)/c$ and
$\widehat q(cs)\,cs_{vp}=\widehat q(s)\,s_{vp}$ for every pixel.  Multiplying a
positive scale by a global constant leaves every calibrated interval unchanged.
\end{lemma}

Both claims follow from positive homogeneity of order statistics.  Dividing
every ratio $r_{vp}/s_{vp}$ by $c$ divides each inner order statistic by $c$,
and therefore divides the outer order statistic of the calibration scores by
$c$ as well; the two factors cancel in the product $\widehat q\,s$.
\hfill$\square$

This one fact underlies four separate conventions in the paper: the
multiplicative non-identifiability of $a$ and $b$, the query-batch mean
normalization of our raw view predictions, the scene-level normalization
applied to every baseline scale map, and the exact cancellation of global
multiplicative bias in Proposition~\ref{prop:decomposition}.

\subsection{Proof of Lemma~\ref{lem:event}}

\begin{lemma}[View event as an order-statistic event]
\label{lem:event}
For a fixed view $v$ and positive scale $s$, the event in
Equation~\eqref{eq:view-event} holds at multiplier $q$ if and only if
$q\ge R_v(s)$.
\end{lemma}

For positive $s_{vp}$,
$Y_{vp}\in C_{vp}(q,s)$ if and only if
$r_{vp}/s_{vp}\le q$.  Consequently the within-view coverage fraction is the
empirical cumulative distribution function of
$\{r_{vp}/s_{vp}:p\in\calP_v\}$.  It first reaches $1-\beta$ at the
$\lceil n_v(1-\beta)\rceil$-th order statistic, which is exactly $R_v(s)$.
\hfill$\square$

\subsection{Proof of Theorem~\ref{thm:validity}}

By Lemma~\ref{lem:event}, the new-view event is equivalent to
$R_V\le\widehat q$.  Under exchangeability, the rank of $R_V$ among the $m+1$
scores is uniform after random tie breaking and conservative without it.  For
$k_\alpha=\lceil(m+1)(1-\alpha)\rceil\le m$,
\begin{equation}
  \Prob\{R_V\le R_{(k_\alpha)}\}
  \ge\frac{k_\alpha}{m+1}\ge1-\alpha.
\end{equation}
If scores have no ties, equality holds in the first relation and
$k_\alpha/(m+1)<1-\alpha+1/(m+1)$.  When $k_\alpha>m$, our convention is
$\widehat q=+\infty$, so the lower bound remains true but the set is
uninformative.  \hfill$\square$

\subsection{Label-free batch construction of the scale}

Theorem~\ref{thm:validity} needs the calibration and test scores to be
exchangeable, but our scales are not fixed functions of one view in isolation:
they are built from the covariates of the whole query batch.  The next
proposition says this is harmless provided the construction ignores the order
of the batch.

\begin{proposition}[Covariate-transductive scale construction]
\label{prop:equivariance}
Let $(Z_1,Y_1),\ldots,(Z_N,Y_N)$ be exchangeable view pairs, with $N>m$, and
let the positive scales be produced jointly from all unlabeled covariates,
\begin{equation}
 (S_1,\ldots,S_N)
 =\Phi(Z_1,\ldots,Z_N;\calD_{\mathrm{source}},\widehat\mu),
 \label{eq:equivariant-scale}
\end{equation}
where $\calD_{\mathrm{source}}$ and the renderer $\widehat\mu$ are frozen
before any target calibration label is observed and $\Phi$ is permutation equivariant,
$\Phi_{\pi(i)}(Z_{\pi(1)},\ldots,Z_{\pi(N)})=\Phi_i(Z_1,\ldots,Z_N)$ for
every permutation $\pi$.  After this construction, draw a uniform calibration
subset of size $m$, independently of the view pairs.  Then the scores
$R_i=R(Z_i,Y_i;S_i)$ are exchangeable, and View-CP satisfies
Theorem~\ref{thm:validity} for a marginal test view from the complement.
\end{proposition}

Permuting the views permutes the covariate vector; by equivariance it permutes
the scale vector the same way, hence the score vector the same way.  A
permutation-equivariant function of an exchangeable sequence is exchangeable.
The uniform split is independent and permutation symmetric, so the $m$
calibration scores and a marginal test score retain the rank symmetry used by
Theorem~\ref{thm:validity}.
\hfill$\square$

Three constructions in this paper satisfy
Equation~\eqref{eq:equivariant-scale}: normalizing our positive raw view
predictions to mean one over the query batch, the scene constant $c_U$ of
Equation~\eqref{eq:baseline-normalization} applied to every baseline map, and
any summary of camera poses or renderer features of the batch.  What matters
is that a single symmetric rule processes calibration and test views together.
Computing separate normalizing constants on the calibration batch and on the
test batch would break the argument.  Our implementation does not do that: all
batch statistics are formed over every query view of a scene, before the
calibration/test split is drawn.

\subsection{Proof of Corollary~\ref{cor:transfer}}

Theorem~\ref{thm:validity} uses only exchangeability of the scores after all
pre-calibration objects are fixed.  It does not require that $s$ estimate a
conditional standard deviation or any true uncertainty function.  Conditional
on $\mathcal F$, the target calibration and test scores are exchangeable by
assumption, so the rank argument is untouched.  Changing $\mathcal F$ may alter
the predictive center, the scale, and thus the score distribution and interval
width, but not the conditional rank guarantee.  \hfill$\square$

\subsection{Proof of Proposition~\ref{prop:oracle}}

By homogeneity of order statistics,
\begin{equation}
 R_v(c\,\widetilde b)
 =Q_{1-\beta,p}\!\left(\frac{r_{vp}}{c\,\widetilde b_{vp}}\right)
 =\frac{1}{c}Q_{1-\beta,p}\!\left(\frac{r_{vp}}{\widetilde b_{vp}}\right)
 =\frac{a_v^\star}{c}.
\end{equation}
Lemma~\ref{lem:event} with $q=1$ says the event holds if and only if
$c\ge a_v^\star$, proving both minimality and the identity.
\hfill$\square$

\subsection{Ranking null space of the view factor}

\begin{lemma}[Ranking null space]
\label{lem:nullspace}
Fix a view $v$ and a positive shape $\widetilde b_v$.  For every $a_v>0$ the
within-view ranking is unchanged,
$\operatorname{rank}_p(a_v\widetilde b_{vp})
=\operatorname{rank}_p(\widetilde b_{vp})$, so any metric $M_v$ that reads the
scale only through that ranking satisfies
$M_v(a_v\widetilde b_v,r_v)=M_v(\widetilde b_v,r_v)$.  At the same time
$R_v(a_v\widetilde b_v)=R_v(\widetilde b_v)/a_v$ by
Proposition~\ref{prop:oracle}.
\end{lemma}

Multiplication by a positive scalar is strictly increasing, so it preserves the
order of $\{\widetilde b_{vp}\}_p$, and per-view AUSE and Spearman correlation
are functions of that order alone.  \hfill$\square$

The two halves of the lemma are the two halves of
Figure~\ref{fig:nullspace}: the view factor is invisible to per-view ranking
diagnostics (Table~\ref{tab:ranking}) and is exactly the quantity that divides
the conformal width (Table~\ref{tab:ablation}).

\subsection{Proof of Theorem~\ref{thm:width}}

Equation~\eqref{eq:ratio-identity} gives calibration scores
$R_v(a_v\widetilde b)=e^{-\delta_v}$.  Their $k_\alpha$-th order
statistic therefore lies in $[e^{-\epsilon},e^{\epsilon}]$.  Since
$\widetilde b$ has mean one in every view,
\begin{equation}
 W_i(a)=\widehat q\,a_i
 =\widehat q a_i^\star e^{\delta_i}
 \in[e^{-2\epsilon}a_i^\star,e^{2\epsilon}a_i^\star].
\end{equation}
For the oracle factor, every calibration score equals one, so
$\widehat q=1$ and $W_i(a^\star)=a_i^\star$.  Dividing proves the result.
\hfill$\square$

\subsection{Proof of Proposition~\ref{prop:decomposition}}

Equation~\eqref{eq:ratio-identity} gives scores $e^{-\delta_v}$.  Because the
exponential is strictly decreasing, the $k_\alpha$-th smallest score is the
$(m-k_\alpha+1)$-th smallest calibration error.  The integer identity
\begin{equation}
 m-\lceil(m+1)(1-\alpha)\rceil+1
 =\lfloor(m+1)\alpha\rfloor=j_\alpha
\end{equation}
therefore gives $\widehat q=e^{-\delta_{(j_\alpha)}}$.  Since
$\widetilde b$ has within-view mean one,
\begin{equation}
 W_i(a)=\widehat q\,a_i
 =a_i^\star e^{\delta_i-\delta_{(j_\alpha)}}.
\end{equation}
The oracle has $W_i(a^\star)=a_i^\star$, proving the pointwise identity.
Averaging numerator and denominator separately yields
Equation~\eqref{eq:aggregate-decomposition}.  \hfill$\square$

\subsection{Proof of Equation~\eqref{eq:realizable-gain-main}}

Fix the calibration and test sets.  By Proposition~\ref{prop:oracle} the oracle
factor gives every calibration view a score of one, so $\widehat q=1$ and its
mean half-width on test view $i$ is $a_i^\star$.  A constant view factor gives
calibration scores $a_v^\star$, hence outer multiplier
$Q_{k_\alpha}(\{a_v^\star:v\in\mathrm{cal}\})$ and that same mean
half-width on every test view.  Taking the ratio of test averages proves
Equation~\eqref{eq:realizable-gain-main}, which therefore also supplies a
direct numerical audit of the implementation.  \hfill$\square$

\section{Implementation and Protocol Details}
\label{app:implementation}

\subsection{Protocol manifest}
\label{app:manifest}

Table~\ref{tab:manifest} lists every scene with its view counts.  Renderer
training views fit the target-scene 3DGS and its residual field.  The remaining
held-out views $V$ are split into $m=\max(11,\lfloor V/3\rfloor)$ calibration
views and $V-m$ test views.  We redraw that split 200 times with seed 0, and
100 times for the native-calibration audit of Table~\ref{tab:validity}.
Repeated splits reuse the same views, so every bootstrap interval, win count
and sign test in the paper resamples \emph{scenes}, never splits.

\begin{table}[h]
\caption{Per-scene manifest.  $V$ is the number of held-out views and $m$ the
calibration size.  Gaussians are counted in the seed-0 model.}
\label{tab:manifest}
\centering
\small
\setlength{\tabcolsep}{4.5pt}
\begin{tabular}{llrrrrr}
\toprule
Scene & Dataset & Train & $V$ & $m$ & Test & Gauss. (k) \\
\midrule
truck     & Tanks \& Temples & 219 & 32 & 11 & 21 & 276 \\
train     & Tanks \& Temples & 263 & 38 & 12 & 26 & 220 \\
drjohnson & Deep Blending    & 230 & 33 & 11 & 22 & 388 \\
playroom  & Deep Blending    & 196 & 29 & 11 & 18 & 246 \\
\midrule
bicycle   & Mip-NeRF~360 & 169 & 25 & 11 & 14 & 858 \\
bonsai    & Mip-NeRF~360 & 255 & 37 & 12 & 25 & 251 \\
counter   & Mip-NeRF~360 & 210 & 30 & 11 & 19 & 197 \\
flowers   & Mip-NeRF~360 & 151 & 22 & 11 & 11 & 662 \\
garden    & Mip-NeRF~360 & 161 & 24 & 11 & 13 & 511 \\
kitchen   & Mip-NeRF~360 & 244 & 35 & 11 & 24 & 284 \\
room      & Mip-NeRF~360 & 272 & 39 & 13 & 26 & 210 \\
stump     & Mip-NeRF~360 & 109 & 16 & 11 &  5 & 629 \\
treehill  & Mip-NeRF~360 & 123 & 18 & 11 &  7 & 804 \\
\bottomrule
\end{tabular}
\end{table}

\paragraph{Source pools.}
Shrinkage selection uses eight source scenes: truck, train, drjohnson,
playroom, bicycle, counter, garden, and room.  The same-family transfer in
Table~\ref{tab:transfer} learns the view regressor on those eight and tests on
the five Mip-NeRF~360 scenes they do not contain, namely bonsai, flowers,
kitchen, stump, and treehill.  The strict holdout uses only truck, train,
drjohnson, and playroom as sources, and tests on all nine Mip-NeRF~360 scenes.
No Mip-NeRF~360 view enters the strict source pool in any role.

\paragraph{Query batch.}
The batch is every held-out view of the target scene, that is all $V$ views of
Table~\ref{tab:manifest}, and it is formed before the calibration/test split is
drawn.  Only unlabeled covariates are read from it: camera poses and
renderer-derived maps.  This is the map $\Phi$ of
Proposition~\ref{prop:equivariance}.

\paragraph{View descriptor.}
The deployed descriptor has 24 entries.  Five are camera geometry
(\texttt{g\_dnn}, \texttt{g\_dk}, \texttt{g\_ang\_min}, \texttt{g\_ang\_k},
\texttt{g\_dens}).  Eight are render statistics (\texttt{r\_alpha} and
\texttt{r\_grad}, each as mean, 10th and 50th percentile, plus
\texttt{r\_depth\_mean} and \texttt{r\_depth\_cv}).  Nine are support
statistics (\texttt{sup\_logE}, \texttt{sup\_ang}, \texttt{sup\_spread}, each
as mean, 10th and 50th percentile).  Two are view-level log-means of spatial
fields (\texttt{f\_gsu\_logmean}, \texttt{f\_train\_logmean}).  A 25th
feature, the log-mean of an out-of-fold residual field, exists in the
source-side schema but is excluded from the deployed schema, because it would
need fold models at target inference.  The two schemas are compared entry by
entry when a fit is loaded, and a mismatch stops the run.

\paragraph{Hyperparameters.}
We select $\lambda_a$ and $\lambda_b$ jointly on the source pool over
$\{0,0.25,0.5,0.75,1\}$, rebuilding the regression target at each $\lambda_b$
so that it always matches the shape that will be deployed.  The ridge penalty
is chosen over $\{0.1,1,10,100,1000\}$ by leave-one-scene-out Spearman
correlation against the oracle factor.  Baseline $\lambda_U$ uses the same
grid and is scored once on the whole pool, rather than by averaging per-scene
minima, because such an average can fall between grid points.  The selected
values are $\lambda_a=1.0$ and $\lambda_b=0.5$ for our scale, $\lambda_U=0.5$
for the 3DGS-U and ensemble maps, and $\lambda_U=0.5$ with $\kappa=64$ for
GAVIS.  The strict-holdout ridge is 100.

\paragraph{Label access.}
Table~\ref{tab:access} states what each group of views may supply.  The last
row is the one that matters for Theorem~\ref{thm:validity}: target test views
contribute camera poses and renderer features to the query batch, and never
contribute an image.

\begin{table}[h]
\caption{What each group of views may supply.  ``Selection'' covers the view
regressor, the ridge penalty, and every shrinkage constant.  Target
calibration images enter only the outer order statistic of
Equation~\eqref{eq:outer-quantile}.}
\label{tab:access}
\centering
\small
\begin{tabular}{lcccc}
\toprule
View group & RGB image & Camera pose & Renderer features & Selection \\
\midrule
Target-scene renderer training & yes & yes & yes & no \\
Source-scene meta-training & yes & yes & yes & yes \\
Target calibration & yes & yes & yes & no \\
Target test & no & yes & yes & no \\
\bottomrule
\end{tabular}
\end{table}

\subsection{Feature construction}

For camera centers $c_v$ and forward axes $h_v$, geometry features include the
nearest and mean five-neighbor distances (normalized by scene extent), minimum
and mean five-neighbor angular differences, and local camera density.  Render
features aggregate alpha, image gradient, and normalized depth.  Support
features aggregate the three per-Gaussian fields in
Equation~\eqref{eq:support}.  Each map contributes its mean and selected lower
quantiles; the spatial field contributes view-level log-mean summaries.  Source
and deployment schemas are checked to exclude any feature that requires
fold-model residuals at target inference.

Three packed auxiliary renders contain: (i) ones, depth, and log exposure;
(ii) support angle, support spread, and the spherical-harmonic spatial field;
and (iii) a training-residual summary.  Together with RGB, this gives four
passes.  Packing is exact because each output channel uses the same weights but
is accumulated independently.

\subsection{Fair normalization of baseline scale maps}

For a baseline map $U_{vp}$, we compute one label-free scene constant over the
unlabeled query batch,
\begin{equation}
 c_U=\frac{1}{|\calV|}\sum_{v\in\calV}\frac{1}{n_v}\sum_p U_{vp},
 \qquad
 s^{U}_{vp}=(1-\lambda_U)+\lambda_U\frac{U_{vp}}{c_U}.
 \label{eq:baseline-normalization}
\end{equation}
This keeps the baseline's cross-view magnitude.  The construction is
permutation equivariant (Proposition~\ref{prop:equivariance}) and uses no image
label, and its global constant is absorbed by the conformal multiplier
(Lemma~\ref{lem:gauge}).  We select $\lambda_U$ on source scenes.
In contrast, our spatial field is normalized inside each view because its
cross-view magnitude is assigned to $a_v$.  Applying that identifiability rule
to a baseline would remove view-level information from its raw scale map.

This choice moves the main-table numbers, so Table~\ref{tab:norm} reports the
whole grid rather than only the selected point.  Per-view normalization divides
each baseline map by its own view mean, which deletes the cross-view magnitude
that the map carries.  Scene-level normalization keeps that magnitude and fixes
only the global unit, which Lemma~\ref{lem:gauge} shows the conformal
multiplier would absorb anyway.  The $\lambda=1$ row is the raw scale with no
shrinkage.  Scene-level normalization is better than per-view normalization for
all three baselines at every $\lambda>0$, so the reported baselines are the
stronger of the two versions, not the weaker.

\begin{table}[h]
\caption{Baseline normalization sensitivity.  Entries are pool-mean relative
width, so lower is better and $\lambda=0$ is the constant scale by definition.
``pv'' normalizes each map inside each view; ``sg'' uses one scene constant.
The last row is the 13-scene saving at the $\lambda$ selected on the pool,
marked $^\ast$.}
\label{tab:norm}
\centering
\small
\begin{tabular}{lcccccc}
\toprule
& \multicolumn{2}{c}{3DGS-U field} & \multicolumn{2}{c}{Ensemble std.}
& \multicolumn{2}{c}{GAVIS vis.} \\
\cmidrule(lr){2-3}\cmidrule(lr){4-5}\cmidrule(lr){6-7}
$\lambda$ & pv & sg & pv & sg & pv & sg \\
\midrule
0.00 & 1.0000 & 1.0000 & 1.0000 & 1.0000 & 1.0000 & 1.0000 \\
0.25 & 0.9326 & 0.8781 & 0.9078 & 0.8097 & 0.9878$^\ast$ & 0.9579 \\
0.50 & 0.9180$^\ast$ & 0.8329$^\ast$ & 0.8968$^\ast$ & 0.7772$^\ast$ & 0.9892 & 0.9416$^\ast$ \\
0.75 & 0.9505 & 0.8413 & 0.9505 & 0.8115 & 1.0063 & 0.9561 \\
1.00 & 1.0831 & 0.9880 & 1.1352 & 0.9521 & 1.0470 & 1.0081 \\
\midrule
Saving (\%) & 9.9 & \textbf{17.4} & 10.5 & \textbf{21.0} & 1.4 & \textbf{4.2} \\
\bottomrule
\end{tabular}
\end{table}

The bold savings are the values reported in Table~\ref{tab:panel-a}.  Had we
used per-view normalization instead, the three baselines would have scored
9.9\%, 10.5\% and 1.4\%, and our 22.1\% would have looked far stronger than it
should.  The gap is largest for the ensemble, whose cross-view magnitude is its
most useful signal.

\subsection{Calibration feasibility}

The smallest calibration size is the first $m$ satisfying
\begin{equation}
  \lceil(m+1)(1-\alpha)\rceil\le m.
\end{equation}
For commonly used levels,
\begin{center}
\begin{tabular}{ccc}
\toprule
$\alpha$ & Target view-event probability & Minimum $m$ \\
\midrule
0.10 & 0.90 & 9 \\
0.05 & 0.95 & 19 \\
\bottomrule
\end{tabular}
\end{center}
At smaller $m$, a finite threshold cannot carry the stated distribution-free
guarantee.  This discreteness must not be hidden by an interpolated quantile.

\section{Additional Experiments and Diagnostics}
\label{app:additional-experiments}

\subsection{Native scores versus conformalized intervals}

Table~\ref{tab:validity} compares three readings of the same uncertainty map.
``Native'' uses the published residual or Gaussian interpretation without a
view-level correction.  Pixel pooling selects one multiplier from all
calibration pixels.  View-CP uses
Equations~\eqref{eq:view-score}--\eqref{eq:outer-quantile}.

\begin{table}[h]
\caption{Coverage averaged over 13 scenes at
$(\alpha,\beta)=(0.1,0.1)$ with $m=\max(11,\lfloor V/3\rfloor)$, the calibration
size used throughout the main results.  Each entry is pooled-pixel / view-event coverage
(\%).  View-CP satisfies the targeted view event; pooled calibration can match
marginal pixels while leaving roughly 40\% of views below target.}
\label{tab:validity}
\centering
\resizebox{\linewidth}{!}{%
\begin{tabular}{lccc}
\toprule
Scale map & Native interpretation & Pixel-pooled CP & View-CP \\
\midrule
Ensemble standard deviation & 30.5 / 0.0 & 89.9 / 59.7 & 94.8 / 92.0 \\
3DGS-U spatial field         & 72.5 / 0.6 & 89.9 / 61.2 & 95.3 / 92.0 \\
Ours: $a\times\widetilde b$ & n/a        & 89.9 / 61.4 & 95.0 / 91.7 \\
\bottomrule
\end{tabular}}
\end{table}

For completeness, Table~\ref{tab:curve} gives the values plotted in
Figure~\ref{fig:validity-efficiency}(a).

\subsection{The view event as a distribution}
\label{app:cdf}

\begin{figure}[h]
  \centering
  \includegraphics[width=0.62\linewidth]{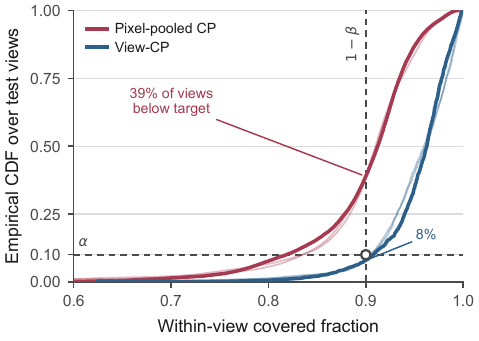}
  \caption{Within-view covered fraction at $(\alpha,\beta)=(0.1,0.1)$, as an
  empirical CDF over 23{,}100 held-out (view, split) pairs from 13 scenes.  The
  bold curves use the 3DGS-U field, the map of
  Figure~\ref{fig:validity-efficiency}(a).  The faint curves behind them are
  the ensemble standard deviation and our scale.  The dashed lines and the open
  circle mark the guarantee: a valid procedure passes below
  $(1-\beta,\alpha)$.}
  \label{fig:cdf}
\end{figure}

Table~\ref{tab:validity} reports two averages.  Figure~\ref{fig:cdf} shows the
distribution behind them.  The $x$ axis is the fraction of a view's pixels that
the interval covers.  The curve is the empirical distribution function over
held-out test views, pooled over the same 13 scenes, the same $m$, and the same
100 splits.

The guarantee is a corner, not a curve.  Theorem~\ref{thm:validity} asks that at
most $\alpha$ of views fall below $1-\beta$, that is $F(1-\beta)\le\alpha$.
Drawing both lines makes that checkable by eye.  View-CP passes under the
corner at 7.9\%.  Pixel-pooled CP misses it at 38.9\%.

At this operating point $\gamma=0.9$, so the matched requirement
$\beta=1-\gamma$ equals the fixed $\beta=0.1$, and nothing in the figure
depends on which convention is used.  That is worth saying, because at other
levels it does matter.  With $\beta$ pinned while $\gamma$ moves, a pooled
multiplier calibrated to a marginal rate below $1-\beta$ cannot satisfy the
view event for arithmetic reasons, and a flat pooled curve would then be forced
rather than observed.  Figure~\ref{fig:validity-efficiency}(a) matches $\beta$
to the level for the same reason.

The three scale maps give almost the same pair of curves.  Pooled CP fails on
38.0--40.0\% of views and View-CP on 7.9--8.3\%.  The split is a property of
the calibration unit, not of the scale map: Theorem~\ref{thm:validity} holds
for each of the three.

\begin{table}[h]
\caption{View-event calibration curve (\%) for the 3DGS-U spatial field, over
the ten scenes of Figure~\ref{fig:validity-efficiency}(a) at $m=21$.  Here
$\gamma=1-\alpha=1-\beta$, so the within-view requirement moves with the level
and pooled calibration is never arithmetically prevented from meeting it.}
\label{tab:curve}
\centering
\begin{tabular}{lrrrrrr}
\toprule
Nominal $\gamma$ & 50 & 60 & 70 & 80 & 90 & 95 \\
\midrule
Pixel-pooled CP & 56.3 & 57.9 & 59.8 & 60.6 & 62.0 & 61.7 \\
View-CP & 50.6 & 63.7 & 72.6 & 82.1 & 90.9 & 95.1 \\
\bottomrule
\end{tabular}
\end{table}

When a baseline publishes a score map rather than a calibrated RGB interval,
we report the two objects separately.  Reading the ensemble map as a Gaussian
standard deviation gives only 30.5\% marginal pixel coverage at nominal 90\%;
in the sweep run of Table~\ref{tab:curve}, View-CP multiplies its native
Gaussian radius by 4.84, while the 3DGS-U residual map starts at 72.5\% pixel
coverage and requires a 2.31$\times$ multiplier.  These
large corrections do not imply poor ranking.  The ensemble has the best AUSE
in Table~\ref{tab:ranking}.  Instead, they show that ranking units are not
automatically interval units.  Constant, visibility, and our normalized scale
have no native standard-deviation interpretation, so we do not manufacture one.

\subsection{GAVIS visibility-field scope and sanity check}
\label{app:gavis}

We import the released degree-two spherical-harmonic visibility representation
from GAVIS, but not its separate uncertainty-aware Bayesian rasterizer.  For a
Gaussian and view, we replace the released last-written-pixel transmittance by
the footprint-weighted average
\begin{equation}
 \overline T_{i,v}=\frac{\sum_p\alpha_{ivp}T_{ivp}}
 {\sum_p\alpha_{ivp}}\in[0,1],
\end{equation}
then alpha-composite visibility in our shared rasterizer.  We select
$\kappa\in\{1,4,16,64\}$ and shrinkage on source scenes; $\kappa=64$ is best,
while the released $\kappa=1$ saturates under our much denser training-view
regime.  The directional field passes its intended sanity check:

\begin{center}
\begin{tabular}{lrrrr}
\toprule
Scene & Training dir. & Test dir. & Random dir. & Opposite dir. \\
\midrule
truck & 0.4332 & 0.4332 & 0.1341 & 0.2324 \\
bicycle & 0.3181 & 0.3220 & 0.0947 & 0.1224 \\
\bottomrule
\end{tabular}
\end{center}
Training directions are 3.2--3.4$\times$ more visible than random directions,
but interleaved test views are as visible as training views.  This supports the
mechanism interpretation in Section~\ref{sec:experiments} without claiming a
full-system reproduction.

\subsection{Calibration-size sweep}
\label{app:calibration-sweep}

For the $m$ sweep, we use the same seven scenes for every setting and hold the
test set at five views so that only calibration size changes.  On stump,
increasing $m$ from 9 to 11 changes saving only from 19.9\% to 20.1\%; the
important cost is crossing the finite-sample feasibility threshold.

\begin{center}
\begin{tabular}{rrrrrr}
\toprule
$m$ & 9 & 11 & 15 & 20 & 25 \\
\midrule
Saving (\%) & 20.7 & 21.4 & 23.0 & 17.1 & 18.1 \\
Event cov. (\%) & 89.6 & 91.4 & 94.0 & 90.5 & 92.4 \\
\bottomrule
\end{tabular}
\end{center}
The non-monotonic saving is predicted by the discrete outer order statistic:
for $m\in\{9,11,15\}$, $k_\alpha=m$ and calibration uses the maximum score;
for $m\in\{20,25\}$ it does not.  More calibration labels buy feasibility and
stability, not monotonic width reduction.

\subsection{Factorization, ranking metrics, and theory audit}
\label{app:factor-diagnostics}

A target built from one deployed model obtains 21.4\% mean saving on the
eight-source pool, matching an 11-training out-of-fold target (20.7\%) at one
eleventh of the source training cost.  The out-of-fold version is better on
6/8 scenes but is 4.9\% worse than constant on playroom, illustrating that a
good average ranking can still have a harmful lower tail.

\begin{table}[h]
\caption{Uncertainty ranking versus interval efficiency over 13 scenes.
AUSE and Spearman are computed separately within each view.}
\label{tab:ranking}
\centering
\begin{tabular}{lccc}
\toprule
Scale & AUSE $\downarrow$ & Spearman $\uparrow$ & Saving (\%) $\uparrow$ \\
\midrule
GAVIS vis. ($\kappa=64$) & 0.4796 & 0.1302 & 4.2 \\
Ensemble std. & \textbf{0.2331} & \textbf{0.4657} & 21.0 \\
Ours: spatial shape $\widetilde b$ & 0.2479 & 0.4604 & 9.9 \\
Ours $a\times\widetilde b$ & 0.2479 & 0.4604 & \textbf{22.1} \\
\bottomrule
\end{tabular}
\end{table}

Finally, we numerically audit Proposition~\ref{prop:decomposition} over 13
scenes and 200 splits.  The three identities in
Equation~\eqref{eq:exact-width} and
Equation~\eqref{eq:aggregate-decomposition} hold to relative error below
$1.1\times10^{-6}$.  After centering the non-identifiable global error, the
test-side factor is 0.975 and the calibration-tail factor is 1.420, yielding
the observed oracle width ratio 1.384.  Hence almost all excess width is paid
for the lower tail of view-factor error.  Here $j_\alpha=1$ in all 13 scenes:
one underestimated calibration view determines the global conformal inflation.

\subsection{Sensitivity to the coverage targets}

\begin{table}[h]
\caption{Sensitivity of saving and view-event coverage.  In the $\beta$ sweep,
the view predictor remains the one trained for $\beta=0.1$, deliberately
testing scale misspecification.}
\label{tab:sensitivity}
\centering
\begin{tabular}{lcccc}
\toprule
Sweep & Setting & Scenes & Saving (\%) & Event cov. / target (\%) \\
\midrule
$\alpha$ & 0.05 & 9 & 24.2 & 95.6 / 95 \\
             & 0.10 & 9 & 21.4 & 91.7 / 90 \\
             & 0.20 & 9 & 13.9 & 84.2 / 80 \\
\midrule
$\beta$  & 0.05 & 13 & 27.6 & 91.9 / 90 \\
             & 0.10 & 13 & 22.0 & 91.7 / 90 \\
             & 0.20 & 13 & 15.4 & 91.6 / 90 \\
\bottomrule
\end{tabular}
\end{table}

\subsection{Cross-backbone transfer}

We replace FastGS density control by vanilla 3DGS densification while keeping
the rasterizer and all uncertainty code fixed.  This produces 5.6--7.7$\times$
more Gaussians at comparable residual level.  The view regressor and shrinkage
parameters remain those learned on FastGS; they are not refit.

\begin{table}[h]
\caption{Saving (\%) and event coverage for the controlled vanilla-densification
backbone.  The full factorization remains better than either factor alone.}
\label{tab:backbone}
\centering
\begin{tabular}{lrrrrr}
\toprule
Scene & $a$ only & $\widetilde b$ only & Full & Oracle & Event cov. \\
\midrule
truck & 5.2 & 7.7 & 12.0 & 31.0 & 91.6 \\
bicycle & 18.7 & 4.6 & 22.8 & 42.1 & 91.3 \\
room & 7.2 & 14.0 & 20.1 & 45.5 & 92.5 \\
\midrule
Mean & 10.4 & 8.8 & \textbf{18.3} & 39.5 & -- \\
\bottomrule
\end{tabular}
\end{table}

\subsection{Rendering latency and exact channel packing}

\begin{table}[h]
\caption{End-to-end query latency on an RTX 4090, including RGB prediction.
Ours uses one target-scene model and four rasterization passes; the ensemble
uses ten models and ten RGB passes.}
\label{tab:latency}
\centering
\begin{tabular}{lcccc}
\toprule
Scene (MP) & Ours (ms) & Ours FPS & Ensemble (ms) & Speedup \\
\midrule
truck (0.53) & 3.57 & 280 & 8.67 & 2.43$\times$ \\
bicycle (1.02) & 4.63 & 216 & 11.86 & 2.56$\times$ \\
room (1.62) & 3.91 & 256 & 9.06 & 2.32$\times$ \\
\bottomrule
\end{tabular}
\end{table}

Channel packing reduces the previous eight-pass implementation to four passes
without changing predictions: the maximum relative difference is zero for the
key alpha and spatial maps and $6.23\times10^{-8}$ over the complete feature
dictionary.  Table~\ref{tab:latency} shows 216--280 FPS across 0.53--1.62 MP.
These numbers are specific to FastGS and this hardware; they are not a
cross-paper claim against differently measured real-time UQ systems.  Building
the GAVIS visibility field is cheaper than our conjugate-gradient residual
field, but its width result in Table~\ref{tab:panel-a} shows that the performance
gap is not caused by query cost.

\subsection{What a calibrated interval looks like}
\label{app:qualitative}

\begin{figure}[h]
  \centering
  \includegraphics[width=\linewidth]{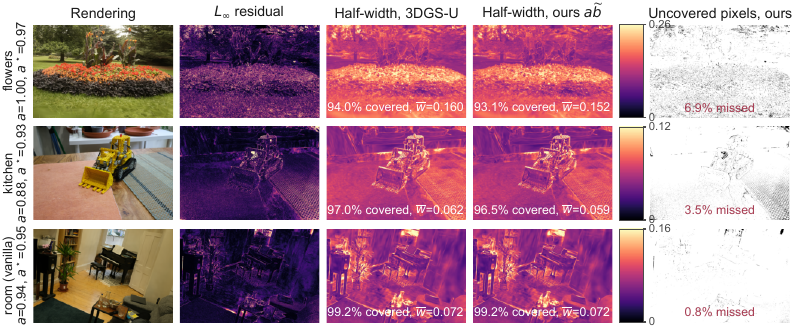}
  \caption{Calibrated intervals on one held-out view per scene, at
  $(\alpha,\beta)=(0.1,0.1)$.  Columns 2--4 of a row share one absolute colour
  scale in RGB units, given by the bar.  Rows do not share it: the three scenes
  sit at 0.26, 0.12 and 0.16 RGB units.  Row labels give the deployed view
  factor $a$ and the oracle $a^\star$, both relative to their batch mean, so
  that the two are comparable.  Inside a single view the two half-width maps
  are close to affine-related, which is Lemma~\ref{lem:nullspace} once more, so
  they are meant to look alike.  The comparison between them is the printed
  mean half-width $\overline w$, not the picture.  The last column marks the
  pixels our interval misses.  All three views clear the 90\% within-view
  target, at 93.1\%, 96.5\% and 99.2\%.}
  \label{fig:qualitative}
\end{figure}

Figure~\ref{fig:qualitative} shows the objects the paper is actually about: an
interval, and the pixels it misses.  We fixed the three scenes in advance to
cover three regimes.  Flowers is spatial-shape dominant, since $\widetilde b$
alone saves 16.8\% there under strict holdout while $a$ alone saves 6.9\%.
Kitchen is our weakest scene, where the view factor alone loses 4.7\% and the
full scale saves only 7.0\%.  Room uses the vanilla-densification backbone,
with the view predictor and both shrinkages carried over from FastGS without
refitting.

Within each scene we show the view of median oracle difficulty $a_v^\star$.
Figure~\ref{fig:nullspace} already shows the two extremes of a scene, so
repeating that rule would show the same thing twice.  The median view answers
the different question of what a typical deployment looks like.  The displayed
view is always excluded from the $m$ calibration views that produce
$\widehat q$.

Three things in the figure are worth stating.  First, the interval is
everywhere much wider than the typical residual.  It has to be: it must cover
90\% of the pixels, so its width is set by the upper tail of the error and not
by the average.  Second, the misses are not spread evenly.  They concentrate on
thin structure and high-frequency texture, which is where a Gaussian primitive
renders an edge as a soft ramp.  Third, on the vanilla-backbone view the two
scales happen to give the same mean half-width.  That is a property of this one
view, not of the scene: over the whole scene the full factorization saves
20.1\% against a constant scale (Table~\ref{tab:backbone}).

\subsection{Per-scene results behind Table~\ref{tab:panel-a}}

Table~\ref{tab:per-scene-main} gives the 13 individual scenes that produce the
means in Table~\ref{tab:panel-a}, together with the two paired differences.
Neither aggregate is driven by a few scenes.  Our scale beats 3DGS-U on 11 of
13, and the two losses are small (kitchen $-6.4$, train $-1.6$).  Against the
ensemble the picture is genuinely mixed, which is why we claim parity rather
than an advantage.  We win on 7 of 13, and the column holds both the largest
win in the table (bonsai, $+24.7$) and the largest loss (drjohnson, $-19.5$).
The ensemble is very strong on drjohnson and room, while our scale is stronger
on bonsai and train.

\begin{table}[h]
\caption{Per-scene width saving (\%) against the constant scale, for all 13
scenes of Table~\ref{tab:panel-a}.  The last two columns are paired
differences on the same scene.  View-event coverage stays within
90.0--93.2\% across all 13 scenes and all five scales.}
\label{tab:per-scene-main}
\centering
\small
\setlength{\tabcolsep}{5pt}
\begin{tabular}{lrrrrrr}
\toprule
Scene & 3DGS-U & GAVIS & Ensemble & Ours & Ours$-$3DGS-U & Ours$-$Ens. \\
\midrule
truck     & 14.6 & $-0.6$ & 10.7 & 17.4 & 2.8 & 6.7 \\
train     & 18.8 & $-2.3$ & 5.7  & 17.2 & $-1.6$ & 11.5 \\
drjohnson & 12.3 & 4.8    & 34.7 & 15.2 & 2.9 & $-19.5$ \\
playroom  & 13.3 & 11.0   & 25.4 & 22.4 & 9.1 & $-3.0$ \\
bicycle   & 15.7 & $-0.2$ & 13.7 & 18.2 & 2.5 & 4.5 \\
bonsai    & 35.3 & $-5.9$ & 18.0 & 42.7 & 7.4 & 24.7 \\
counter   & 18.3 & 1.9    & 20.4 & 24.0 & 5.7 & 3.6 \\
flowers   & 19.4 & 6.7    & 20.5 & 21.0 & 1.6 & 0.5 \\
garden    & 20.8 & 8.9    & 28.2 & 21.8 & 1.1 & $-6.4$ \\
kitchen   & 13.5 & $-6.3$ & 4.9  & 7.1  & $-6.4$ & 2.2 \\
room      & 19.9 & 23.2   & 39.4 & 35.1 & 15.2 & $-4.3$ \\
stump     & 11.3 & 4.7    & 22.0 & 19.3 & 8.0 & $-2.7$ \\
treehill  & 13.0 & 8.8    & 29.1 & 26.2 & 13.1 & $-2.9$ \\
\midrule
Mean      & 17.4 & 4.2 & 21.0 & \textbf{22.1} & 4.7 & 1.2 \\
Wins      & \multicolumn{4}{l}{} & 11/13 & 7/13 \\
Sign $p$  & \multicolumn{4}{l}{} & 0.0225 & 1.0 \\
\bottomrule
\end{tabular}
\end{table}

\subsection{Cross-family transfer and predictive-center controls}

Table~\ref{tab:transfer} reports the aggregate transfer results omitted from
the main text.  In the same-family setting, the view regressor is learned on
eight source scenes and tested on five unseen Mip-NeRF~360 scenes.  In strict
family holdout it is learned only on two Tanks \& Temples and two Deep Blending
scenes, then used without refitting on all nine Mip-NeRF~360 scenes.  All
hyperparameters remain source-selected.

\begin{table}[h]
\caption{Cross-scene transfer with a shared single-model center.  Savings are
relative to a constant scale.  Gap and wins compare VSCP with the
source-selected ensemble scale.}
\label{tab:transfer}
\centering
\begin{tabular}{lccccc}
\toprule
Transfer & Ours (VSCP) & Ensemble scale & Gap [95\%] & Wins & Sign $p$ \\
\midrule
Same family (5 scenes) & 21.8 & 18.2 & 3.6 [-1.9, 12.5] & 3/5 & 1.0 \\
\rowcolor{OursRow}
Strict family holdout (9) & \key{20.7} & 20.0 & 0.7 [-4.2, 6.5] & 4/9 & 1.0 \\
\bottomrule
\end{tabular}
\end{table}

A four-scene source pool of the same size that includes Mip-NeRF~360 gives a
21.9\% saving, only 1.2 points above strict holdout.  Thus the observed domain
shift has little efficiency cost, while target-scene View-CP remains the safety
layer.  The strict predictor beats the constant scale on all nine target scenes
($p=0.0039$) and matches an ensemble scale that requires ten models trained on
each target scene.  This comparison concerns how the view-difficulty signal is
obtained, not a claimed width advantage over the ensemble.

The conclusion is unchanged when both scales use the ensemble mean as their
predictive center.  Under strict holdout, VSCP saves 20.0\% and the native
ensemble scale 18.1\%; the paired gap is 1.9 points [-3.8, 8.3], with 5/9 wins
and $p=1.0$.  In the same-family setting the gap is 4.6 points [-1.8, 14.2],
with 4/5 wins and $p=0.375$.  When ten models are available, the ensemble mean
and the VSCP scale can therefore be combined.

\subsection{Scene-level strict-holdout results}

\begin{table}[t!]
\caption{Per-scene width saving (\%) in strict leave-Mip-NeRF-360-out
meta-training. The source pool contains only Tanks \& Temples and Deep
Blending.}
\label{tab:per-scene}
\centering
\small
\setlength{\tabcolsep}{4pt}
\begin{tabular}{lrrrrr}
\toprule
Scene & $a$ only & $\widetilde b$ only & Full & Ensemble scale & Oracle \\
\midrule
bicycle  & 14.8 & 4.7  & 20.0 & 13.5 & 40.3 \\
garden   & 10.7 & 15.1 & 19.9 & 25.9 & 37.7 \\
room     & 8.1  & 13.7 & 20.4 & 29.9 & 42.7 \\
counter  & 12.7 & 12.0 & 23.6 & 19.9 & 37.1 \\
flowers  & 6.9  & 16.8 & 19.7 & 19.9 & 29.1 \\
stump    & 11.6 & 5.8  & 18.3 & 21.7 & 36.6 \\
treehill & 13.9 & 6.3  & 22.5 & 28.6 & 40.1 \\
bonsai   & 22.2 & 21.0 & 35.1 & 15.6 & 50.9 \\
kitchen  & -4.7 & 10.6 & 7.0  & 5.0  & 36.5 \\
\midrule
Mean     & 10.7 & 11.8 & \textbf{20.7} & 20.0 & 39.0 \\
\bottomrule
\end{tabular}
\end{table}

The full method beats the constant scale on all nine scenes, with exact
two-sided sign-test $p=0.0039$.  That is the smallest value attainable at
$n=9$, so it records a clean sweep of directions and not an effect size.  The
full method beats the ensemble scale on 4/9 scenes; the paired mean gap is 0.7
points [-4.2, 6.5] and $p=1.0$.  In the five-scene same-family test it beats
the constant on 5/5 scenes, where the floor is $p=0.0625$.  Scene count, rather
than repeated split count, limits all of these tests.

\section{Limitations}
\label{sec:limitations}

The guarantee is marginal over exchangeable views, not conditional coverage for
every camera pose, and ordered camera paths can break exchangeability.  The
method needs labeled target-scene views that are held out.  Using them to train
3DGS instead would need a different calibration argument.  Only one held-out
family is large enough to be useful, and family is confounded with bounded
versus unbounded capture.  The 13-scene results use FastGS.  The
vanilla-densification study covers three scenes and changes only the density
rule, so it is not the official vanilla implementation.  Our GAVIS row
evaluates the released visibility field under a shared rasterizer, not its
Bayesian uncertainty head.  Systems such as Horseshoe Splatting change the
center and the training objective, so we discuss them rather than place them in
Table~\ref{tab:panel-a}.  Finally, our sets are axis-aligned RGB boxes with a
shared radius.  Richer color geometry, dynamic scenes, and time-correlated
calibration units are out of scope.

\end{document}